\documentclass[letterpaper, 10 pt, journal, twoside]{IEEEtran}
\usepackage{amsmath,amsfonts}
\usepackage{algorithmic}
\usepackage{algorithm}
\usepackage{array}
\usepackage[caption=false,font=normalsize,labelfont=sf,textfont=sf]{subfig}
\usepackage{textcomp}
\usepackage{stfloats}
\usepackage{url}
\usepackage{verbatim}
\usepackage{graphicx}
\usepackage{cite}
\usepackage{hyperref} 
\usepackage{epstopdf}
\usepackage{epsfig}  
\usepackage{stfloats}
\usepackage{multirow}
\usepackage{booktabs}
\usepackage{setspace}
\usepackage{color}
\begin{document}

\title{
        \normalsize
        {\textcolor{blue}{This work has been submitted to the IEEE Robotics and Automation Letters\\ (RA-L) for possible publication. Copyright may be transferred without notice, \\after which this version may no longer be accessible}}
        \\[10pt]
        \LARGE
        VG-Swarm: A Vision-based Gene Regulation Network for UAVs Swarm Behavior Emergence}

\author{Yuwei Cai$^{1}$, Huanlin Li$^{1}$, Juncao Hong$^{1}$, Peng Xu$^{1}$, Hui Cheng$^{2}$, Xiaomin Zhu$^{3}$
\\Bingliang Hu$^{4}$, Zhifeng Hao$^{1}$, Zhun Fan$^{1,*}$ % <-this % stops a space
\thanks{This work was supported in part by National Key R\&D Program of China (2021ZD0111502), and in part by the National Natural Science Foundation of China under Grant 62176147. (\textit{Yuwei Cai and Huanlin Li are co-first authors.}) (\textit{Corresponding authors: Zhun Fan.})}% <-this % stops a space
\thanks{$^{1}$Yuwei Cai, Huanlin Li, Juncao Hong, Peng Xu, Zhifeng Hao, and Zhun Fan are with Key Lab of Digital Signal and Image Processing of Guangdong Province, College of Engineering, University of Shantou, Guangdong, China. E-mails: {\tt\footnotesize zfan@stu.edu.cn, 21hlli@stu.edu.cn}}
\thanks{$^{2}$Hui Cheng is with School of Computer Science and Engineering, University of Sun Yat-Sen, Guangdong, China. E-mails: {\tt\footnotesize chengh9@mail.sysu.edu.cn}}
\thanks{$^{3}$Xiaomin Zhu is with College of Systems Engineering National University of Defense Technology Changsha, Hunan, China. E-mails: {\tt\footnotesize xmzhu@nudt.edu.cn}}
\thanks{$^{4}$Bingliang Hu is with Xi'an Institute of Optics and Precision Mechanics, Shanxi, China. E-mails: {\tt\footnotesize hbl@opt.ac.cn}}
}

%\markboth{IEEE Robotics and Automation Letters. Preprint Version. Accepted December, 2022}
%{LI \MakeLowercase{\textit{et al.}}: VG-Swarm: Vision-based UAVs Swarm} 

\maketitle

%%%%%%%%%%%%%%%%%%%%%%%%%%%%%%%%%%%%%%%%%%%%%%%%%%%%%%%%%%%%%%%%%%%%%%%%%%%%%%%%
\begin{abstract}

Unmanned Aerial Vehicles (UAVs) dynamic encirclement is an emerging field with great potential. Researchers often get inspiration from biological systems, either from macro-world like fish schools or bird flocks etc, or from micro-world like gene regulatory networks (GRN). However, most swarm control algorithms rely on centralized control, global information acquisition, and communications among neighboring agents. In this work, we propose a distributed swarm control method based purely on vision and GRN without any direct communications, in which swarm agents of e.g. UAVs can generate an entrapping pattern to encircle an escaping target of UAV based purely on their installed omnidirectional vision sensors. A finite-state-machine (FSM) describing the behavioral model of each drone is also designed so that a swarm of drones can accomplish searching and entrapping of the target collectively in an integrated way. We verify the effectiveness and efficiency of the proposed method in various simulation and real-world experiments.

\end{abstract}

%%%%%%%%%%%%%%%%%%%%%%%%%%%%%%%%%%%%%%%%%%%%%%%%%%%%%%%%%%%%%%%%%%%%%%%%%%%%%%%%
\section{INTRODUCTION}

Recently, the research of swarm UAVs has attracted increasing attention from researchers because of its wide applicability, such as environment exploration \cite{2019Minimal}, target tracking and entrapping \cite{brinon2019circular}, flocking \cite{vasarhelyi2018optimized}, autonomous search and rescue \cite{macwan2014multirobot}, task allocation \cite{gao2022uav} etc, among which a special class of applications includes entrapping and capturing enemy targets, or protecting targets by using multiple UAVs \cite{hafez2014uavs}. 

The main problem to be addressed in this type of application is how to control a group of agents to generate adaptive patterns and entrap specific targets. The mainstream algorithms for pattern generation can be divided into several categories according to inspiration sources, movement constraints, or control methodologies \cite{jin2012hierarchical}, \cite{navarro2013survey}. Actually, our natural world provides many inspiring cases of adaptive pattern generation in self-organizing biological systems, such as the emergence of animals swarm behaviors in flocks, herds, and schools, often leading to interesting pattern formation \cite{reynolds1987flocks}. In the field of micro-organisms, the genome of a cell encodes the interaction among genes, morphogens, and proteins. In a multicellular organism, most cells have the same genome, but the behaviors and the functionalities of each cell are varied and decided by the activation or deactivation of different genes. Regulating gene expression can trigger cell activities including cell differentiation, division, protein production and migration, \cite{west2003developmental}. The process of producing proteins is governed by a gene regulatory network (GRN) \cite{de2002modeling}. 

\begin{figure}[t]
        \centering
        \includegraphics[width=8.5cm]{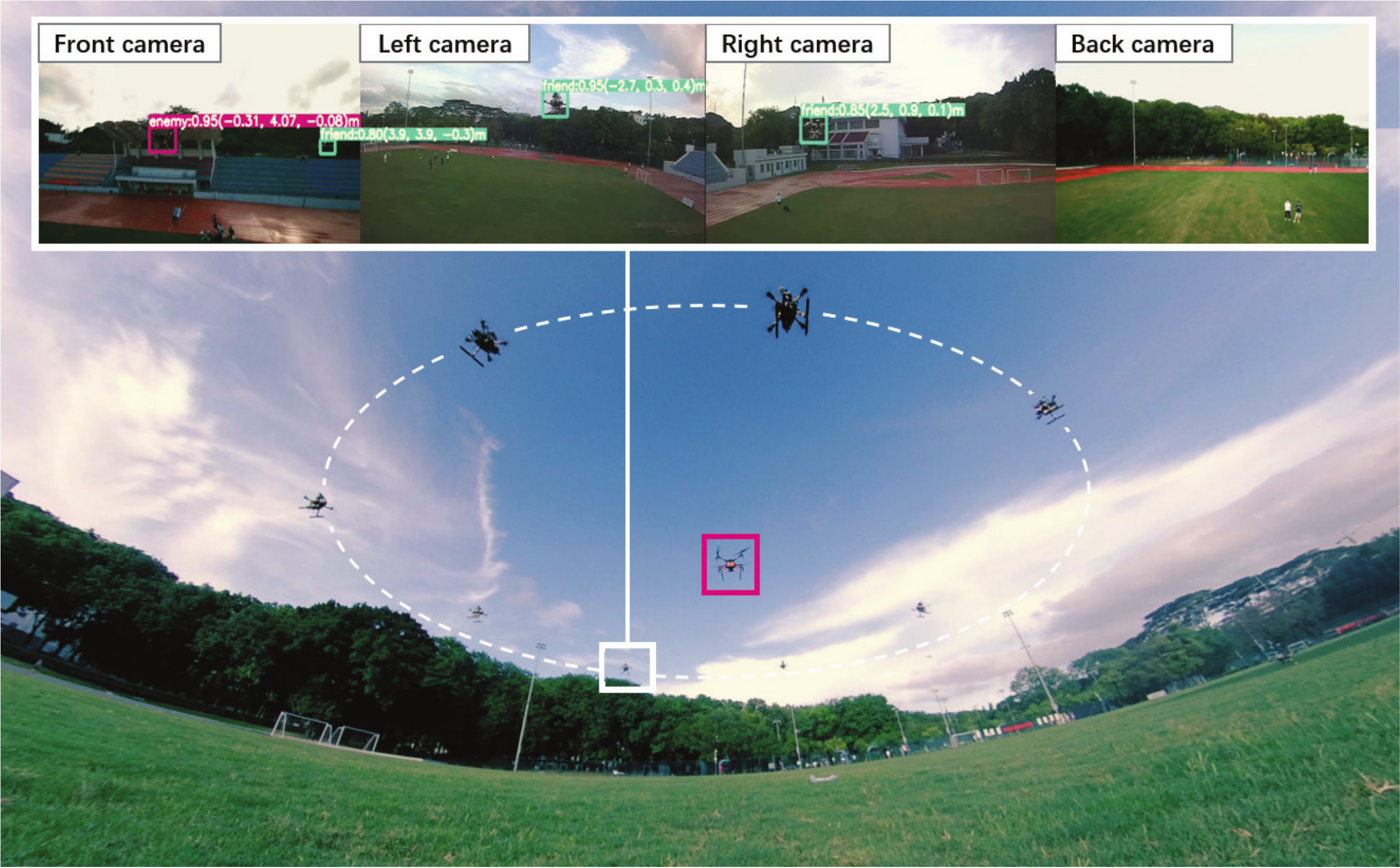}
        \caption{The photo of target entrapping pattern formation for UAVs based on V-GRN in a real world experiment. Each UAV detects and estimates the positions of targets, obstacles and neighbors through omnidirectional images, taking the UAV at the bottom of the photo as an example.}
        \label{fig1}
\end{figure}

In recent years, researchers began to apply mechanisms of GRN in designing control methods for swarm robots \cite{oh2017bio}. Taylor et al. \cite{taylor2007pattern} proposed a GRN-CAM control method by fusing a cell adhesion model (CAM) and a GRN model, which allows the spatial and temporal differentiation of protein expression across swarm robots treated as artificial cells with virtual membranes and artificial cell adhesion molecules. Guo et al. \cite{guo2009cellular} applied a GRN model to generate adaptive patterns for swarm robots. Each robot has two genes that can produce two kinds of proteins to control its movements in x and y directions respectively. These proteins drive the robots to move within the boundary of a predefined shape. Besides, some of the robot motion parameters are optimized by an evolutionary algorithm. However, this method needs to use global coordinates, which are difficult to obtain by robots in many real-world applications. To solve this problem, Guo et al. \cite{guo2011swarm} proposed a method based on GRN that selects a reference robot and sets a local coordinate system. Each robot participates in forming a pattern of an uneven B-spline shape through the local coordinate. Based on the local coordinate system, Jin et al. \cite{jin2012hierarchical} proposed a hierarchical gene regulatory network (H-GRN) for adaptive pattern generation to entrap targets in dynamic environments. Then, the H-GRN model was expanded to achieve area coverage by Oh and Jin et al. \cite{oh2014adaptive}. To deal with the problem of merging and separating patterns in the process of multi-target entrapment and the issue of obstacle avoidance, Oh et al. \cite{oh2014evolving} proposed an EH-GRN structure that uses obstacles and targets as the inputs of an evolving GRN to generate a morphogen gradient space. Similarly, Meng and Guo \cite{meng2012evolving} introduced an evolving GRN method to adjust the structure and parameters of GRN. Lately, Wu et al. \cite{wu2019cooperation} proposed a C-GRN model capable of peer collaboration, which allows agents to discover and reinforce the weak parts of the formed pattern. Yuan et al. \cite{yuan2019th} combined tracking-based H-GRN and the leader-follower model in designing a TH-GRN model that can be used in more complex application environments.

% However, The above GRN models can not solve the problems of limited individual perception range, low efficiency on obstacle avoidance, unknown map information and unable to use centralized controller on agents swarm.
Different from application scenarios in daily life when communications are normally available, swarm robots also need to work in environments when communications are not available, disabled, or largely depressed. In these cases, each agent needs to have the ability to perform the task independently based purely on its own installed sensors, e.g. vision sensors, without receiving any direct communication.

In fact, vision is the main sensory mode of animal groups that allows collective motion \cite{strandburg2013visual}. Renaud Bastien et al. \cite{bastien2020model} demonstrated that organized collective behavior of multi-agent systems can be generated by a vision-based collective behavior model. Fabian Schilling et al \cite{schilling2021vision} carried multiple cameras on a UAV to obtain omnidirectional images and proposed a target tracking algorithm to achieve vision-based UAV flocking. Inspired by the above works, we propose a vision-based GRN model for target entrapment, which opens up a new line of the research on swarm robots for communication-free applications. 
The main contributions of this paper are summarized as follows: 
\begin{itemize}
\item We propose a vision-based gene regulation network (V-GRN) model for communication-free UAVs, which only relies on visual inputs to ensure that swarm UAVs can entrap the targets successfully.
\end{itemize}
\begin{itemize}
\item We improve the detection performance of the original YOLOv5s \cite{glenn_jocher_2021_4679653} algorithm and propose a monocular position estimation algorithm to enable UAVs to better obtain the object positions without relying on communication.
\end{itemize}
\begin{itemize}
\item The behavioral model of each UAV is carefully designed and described in an FSM model so that the swarm UAVs can accomplish searching and entrapping tasks in an integrated way.  
\end{itemize}

Notably, a conference version of this paper appeared in \cite{cai2021behavior} in which we proposed a simplified gene regulation network model for entrapping the targets, validated with experiments using Epuck mobile robots in the 2D indoor workspace. However, this work does not adapt to communicate-free outdoor scenarios, partly due to Epuck's limited sensing capability. This paper addresses this issue by using omnidirectional vision and extending the work so that the improved method can be applied to a swarm of UAVs in the 3D outdoor workspace.

\section{METHOD}

Fig. 2 illustrates the overall framework of the proposed method, which consists of four major parts as follows: perception strategy, object position estimation, a vision-based GRN model (V-GRN), and design of a behavioral model of the swarm of UAVs.
\begin{figure}[t]
        \centering
        \includegraphics[width=8.3cm]{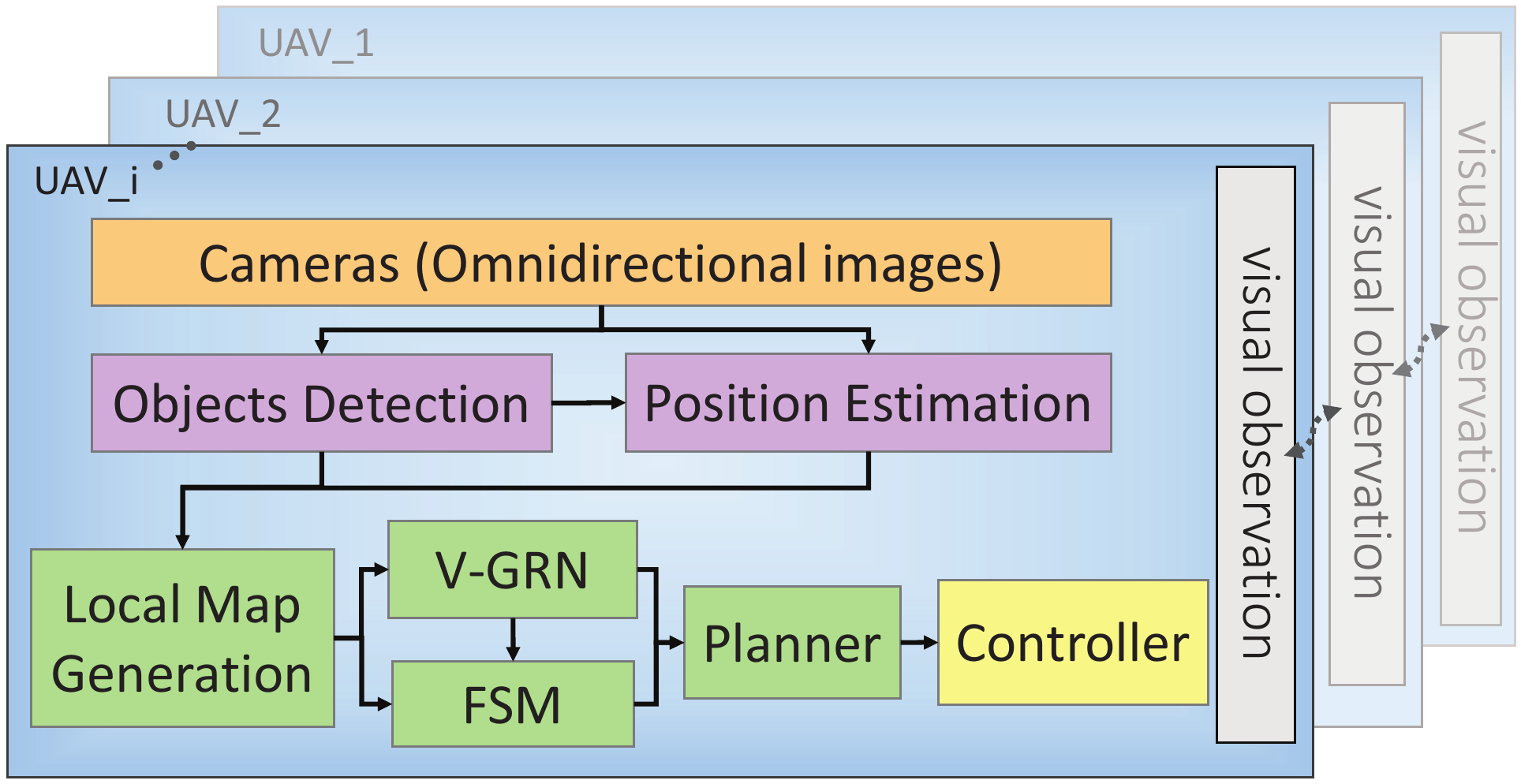}
        \caption{\textbf{The system architecture of the proposed VG-Swarm}. The input, omnidirectional images captured from four cameras, is used for object detection and position estimation. In the central part, a local map records the position information of the observed objects, and V-GRN generates target entrapping patterns using the environmental information provided by the local map. Then the planner generates the command to control the agent's movement according to the FSM model and the calculated data from V-GRN.}
        \label{fig2}
\end{figure}
% The proposed approach for swarm behaviors emergence of communication-free UAVs can be devided into the following parts: perception, swarm strategy, and controller (Fig. 2)

\subsection{Perception Strategy}

\begin{figure*}[t]
        \centering
        \includegraphics[width=16cm]{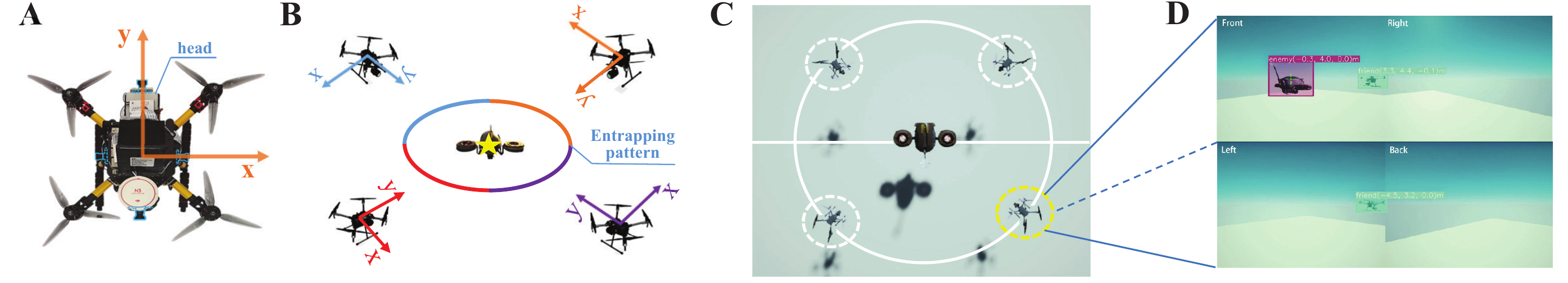}
        \caption{\textbf{The schematic diagram illustrating the local coordinate system, generation of entrapping patterns and its corresponding experimental implementation}. The UAV establishes a right-handed coordinate system with the direction of its own nose as the y-axis. Part B illustrate the entrapping pattern formed by four local patterns. In part C, the final entrapping pattern of the four UAVs is formed in the experiment. Part D shows the visual perception of the UAV in the lower right corner of part C.}
        \label{fig3}
\end{figure*}

Recently YOLO series have been widely used in object detection for their considerable accuracy and fast inference speed. However, directly applying YOLO models to tackle object detection tasks in drone-captured scenarios may not work well due to the usually small sizes of objects detected by drones and the limitation of onboard computing power. In this paper, to further improve the detection performance of small objects on onboard devices, we made some modifications to the original YOLOv5s network architecture \cite{glenn_jocher_2021_4679653}.

In particular, we found that some prediction heads in the original YOLOv5 network are generated from high-level, low-resolution feature maps, which are less sensitive to tiny objects. Therefore, in the neck part, we added one more bottom-up and top-down path and replaced a prediction head for processing the lower-level, higher-resolution feature map. By doing so, the detection performance of the model for tiny objects is slightly improved, but with the number of parameters and calculation time increased. Moreover, We observed that despite the use of a larger feature map, feature diversity of tiny objects is still poor as image details become blurred, and many layer channels contribute little to detect tiny objects. To reduce the parameters, we reduced the number of convolution kernels to around 40\% of the YOLOv5s network.

However, even with the lightweight model, it is still difficult to detect images from multiple cameras in real-time on embedded edge computing devices. To address this issue, we take the following steps: 
\begin{itemize}
\item Using the TensorRT engine to optimize the inference speed of our custom model, and preparing data for the detector using multi-processing.
\end{itemize}
\begin{itemize}
\item Stitching multiple view frames of the same resolution into one, so the detector only needs to detect the combined frame once to get all the visual information.
\end{itemize}
\subsection{Object position estimation}
In this paper, we obtain object position by extending the aforementioned method of online object detection, which combines the detection results and the camera field of view (FOV) information. It is found that there is a certain relationship between the distance of the object and its detected bounding box, which can be well fitted by a power function. Inspired by the work \cite{griffin2021depth}, we use the monocular camera to estimate the relative position of objects from their detected bounding boxes only. 

Furthermore, building on the work of \cite{griffin2021depth}, we replace $\frac{h_i}{h_j}$ with $\sqrt{\frac{s_i}{s_j}}$ and use the optical expansion of the bounding box to obtain the depth Z using two observations at the time $i$, and $j$.
\begin{equation}
\begin{split}
Z_i = \frac{C_{Z_j} - C_{Z_i}}{1 - \sqrt{\frac{s_i}{s_j}}}
\end{split}
\end{equation}

Where $C_{Z_j} - C_{Z_i}$ is the relative camera motion of the two observation points in the depth direction, and $s_i$, $s_j$ are the areas of the bounding boxes at the time $i$ and $j$.

Then we use this formula to fit a distance estimation power function. Furthermore, in order to achieve a better fitting result for distance estimation, we use a Kalman filter to smooth the estimated depth data.
\begin{equation}
\begin{split}
D = \alpha \cdot A^\beta
\end{split}
\end{equation}

Where D is the estimated object distance, A represents the pixel area of the bounding box, $\alpha$ and $\beta$ are the coefficients that need to be fitted.

Finally, the components $X$, $Y$ and $Z$ are calculated from the estimated distance and the camera FOV of horizontal and vertical ($FOV_h$, $FOV_v$) information with reference to the local coordinate system (as shown in Fig. 3) to complete the relative position estimation.
\begin{equation}
\begin{split}
&X = \frac{2x \cdot D \cdot \sin \frac{FOV_h}{2}}{W}  \\
&Y = \sqrt{D^2 - X^2}    \\
&Z = \frac{2y \cdot D \cdot \sin \frac{FOV_v}{2}}{H}
\end{split}
\end{equation}

Where $x$ , $y$ are pixel values that the center of the boundary box deviates from the center of the image, and $W$, $H$ are the width and height pixel values of the image, respectively. 

\subsection{Local map generation}
Local map is used for short-term data association to match map elements (equation (4)) obtained during the last few seconds. The proposed detector can detect objects well in most cases, therefore a local map can be generated in the following procedure: an individual drone first estimates the relative spatial positions of detected objects, then saves the information, including the positions of the detected targets, neighboring UAVs, and obstacles to its local map. 
\begin{equation}
\begin{split}
object := [X, Y, Z, x, y, W, H, T, N]
\end{split}
\end{equation}

The local map records the position information of the object, the object type $(T)$, and the lost threshold $(N)$, where N is used to remove the unobserved objects. We believe that with a sufficiently fast visual perception, the bounding boxes of the same object in the context have a positive intersection over union(IoU). And we use it for position association to update the local map.

\subsection{Vision-based Gene Regulatory Network}

Gene Regulatory Network (GRN) is defined as the network formed by the interaction between genes in cells (or in a specific genome). GRN is the mechanism of controlling gene expression in organisms. When genes are expressed, mRNA stored in the genome is transcribed and translated into proteins. Some of these proteins (such as transcription factors) can regulate themselves or the expression of other genes. Therefore, protein production is regulated and controlled by genes, which forms a complex interactive gene network. The diffusion of proteins generates a concentration field in the cytoplasm that in turn activates or inhibits the expression of other genes \cite{jin2010morphogenetic}. Inspired by this, we designed a control approach for a swarm of UAVs based on GRN. We use the concentration field to represent the task space, with each point in the space having a concentration value, which depends on the local map. 

Compared with traditional GRN models, the proposed V-GRN model (Fig. 4) removes the direct communication part of agents but uses a new vision-based method to obtain information from the environments. 

In the upper layer, proteins $p_1$, $p_2$,and $p_3$ are sensory proteins, which read the positions of the detected targets, obstacles, and neighboring UAVs, respectively. Taking these positions as the inputs, a concentration field is formed in the task space, based on which the target entrapping patterns can be generated. The adaptive pattern generation process is actually equivalent to obtaining a contour closest to the target that at the same time satisfies the following two conditions: 
\begin{itemize}
\item There is an approximately equal number of concentration points in the four quadrants around the target.
\item The minimum distance between the contour line and the target should be greater than a safe distance, which is often predefined by the user.     
\end{itemize}

\begin{figure}[t]
        \centering
        \includegraphics[width=8cm]{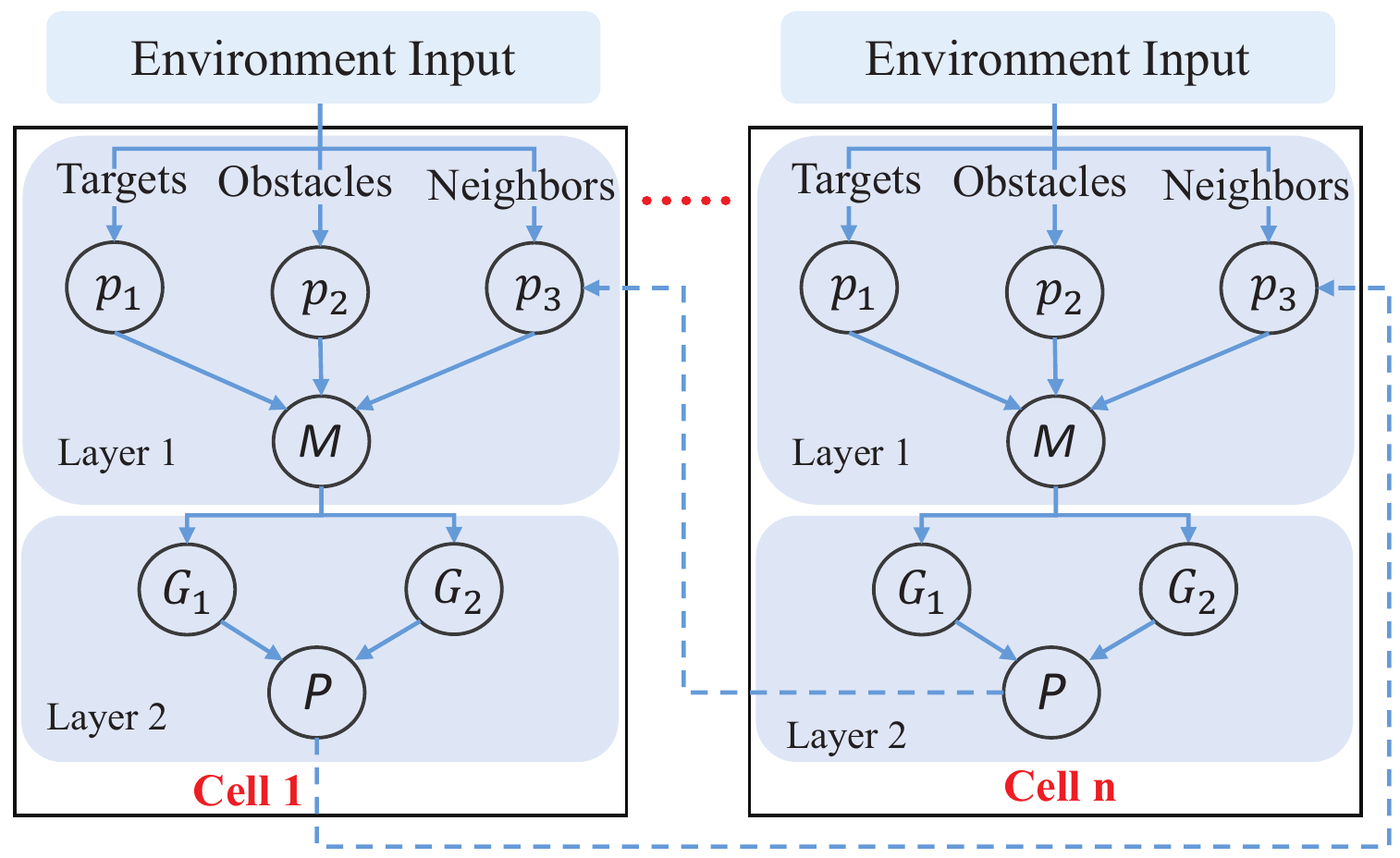}
        \caption{A diagram of the vision-based gene regulatory network (V-GRN). A cell represents an individual UAV, which consists of an upper layer and a lower layer. In the upper layer, $p_1$, $p_2$, and $p_3$ are sensory proteins that can be regulated by external environmental inputs from the targets, obstacles, and neighbors, respectively. $M$ is a protein that fuses the concentration field from proteins $p_1$, $p_2$, $p_3$, and affects the production of proteins $G_1$ and $G_2$, which are actuating proteins in the bottom layer representing control output of orientation and velocity value, respectively. They both affect the production of protein $P$, which ultimately determines the dynamic position of the UAV. In a system, protein $P$ of one cell affects the gene expression of other neighboring cells.}
        \label{fig4}
\end{figure}

In the process of pattern generation, the UAV that discovers the targets first generates the concentration fields about the targets through equations (5) and (9). In the cytoplasm, the diffusion of protein is basically a spatio-temporal process, so the computing formula of concentration includes the derivative term of time.
\begin{equation}
\begin{split}
        \frac{dT_i}{dt}=\bigtriangledown^2T_i + \gamma_i - T_i 
\end{split}
\end{equation}
\begin{equation}
\begin{split}
        \frac{dO_j}{dt}=\bigtriangledown^2O_j + \beta_j - O_j 
\end{split}
\end{equation}
\begin{equation}
\begin{split}
        \frac{dN_m}{dt}=\bigtriangledown^2N_m + \eta_m - N_m 
\end{split}
\end{equation}
% \begin{equation}
% \begin{split}
%         T, O, N = \mathop \sum \limits_{i = 1}^{{n_t}} {T_i}, \mathop \sum \limits_{j = 1}^{{n_o}} {O_j}, \mathop \sum \limits_{m = 1}^{{n_n}} {N_m}    
% \end{split}
% \end{equation}
% \begin{equation}
% \begin{split}
%         O =   
% \end{split}
% \end{equation}
% \begin{equation}
% \begin{split}
        % N = 
% \end{split}
% \end{equation}
% \begin{equation}
% \begin{split}
%         \frac{{dM}}{{dt}} =  - M + sig&({1 - T^2,\theta ,k}) + sig({O^2,\theta ,k}) \\
%         & + sig({N^2,\theta ,k}) 
% \end{split}
% \end{equation}
\begin{equation}
\begin{split}
        \frac{{dM}}{{dt}} =  - M + sig&({1 - (\mathop \sum \limits_{i = 1}^{{n_t}} {T_i})^2,\theta ,k}) + sig({(\mathop \sum \limits_{j = 1}^{{n_o}} {O_j})^2,\theta ,k}) \\
        & + sig({(\mathop \sum \limits_{m = 1}^{{n_n}} {N_m})^2,\theta ,k}) 
\end{split}
\end{equation}
\begin{equation}
\begin{split}
        sig({x,\theta,k}) = \frac{1}{{1 + {e^{ - k( {x - \theta })}}}}  
\end{split}
\end{equation}

Where $\gamma_i$, $\beta_j$, and $\eta_m$ stand for the positions of the $i^{th}$ targets, $j^{th}$ obstacles, and $m^{th}$ neighboring UAVs, respectively. $\bigtriangledown^2$ is a Laplace operator, which is defined as the second derivative of $T_i$, $O_j$, and $N_m$. A resulting concentration field $M$ is obtained according to equation (8) and used to generate the entrapping patterns, in which $\theta$ and $k$ are regulation parameters. 
% $T_i$, $O_j$, and $N_m$ represent the protein concentrations formed by the $i^th$ targets, $j^th$ obstacles, and $m^th$ neighboring UAVs respectively.
Fig. 5 provides an illustrative example assuming that there is only one target in the environment and only one obstacle(or neighboring UAV) around it, in which the generation process of the concentration field is given. The resulting concentration field is the fusion of the above concentration fields. It is notable that the concentration generated by the target now has an opposite sign compared with the concentration generated by the obstacle (as shown in Fig. 5(c)). The closer the location is to the target, the lower the concentration value is in this location, and the opposite is true near the obstacle.

\begin{figure}[t]
        \centering
        \includegraphics[width=8cm]{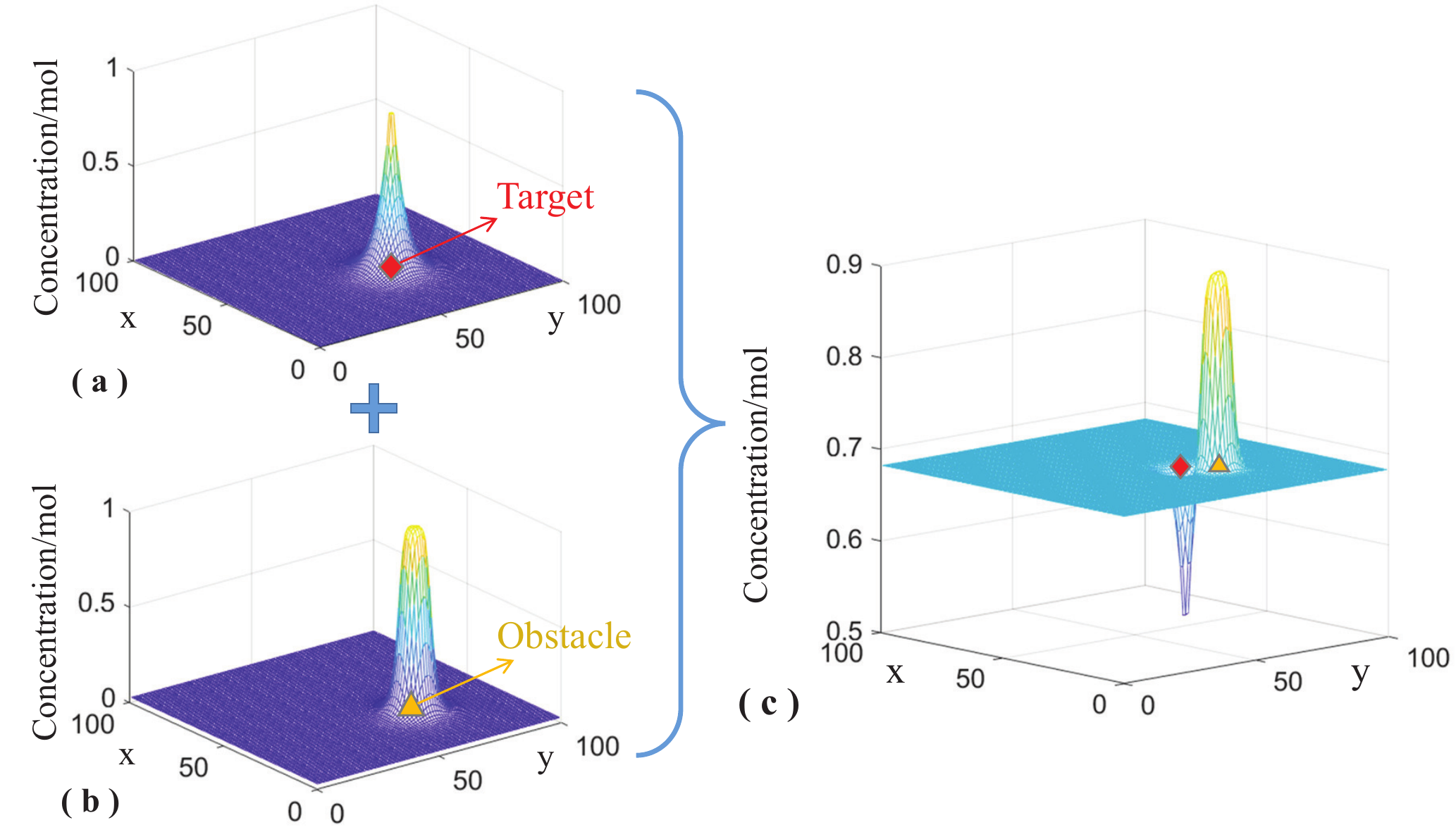}
        \caption{The concentration fields diagram. (a) is the concentration field formed by a target, (b) is the concentration field formed by a obstacle or neighboring UAV, and (c) is the comprehensive concentration field.}
        \label{fig5}
\end{figure}

After generating the concentration space, we take the contour with the same concentration value as the target entrapping pattern, which is the output of the upper layer of V-GRN. The local entrapping pattern generated by each UAV can be merged to form a global entrapping pattern(as shown in Fig. 3(B)). 

In the bottom layer, the actuating proteins $G_1$ and $G_2$ are used to control the movement of the UAV.  We use the method of sampling and selecting the minimum concentration points to determine the vector $\vec{v}$ of the next movement of UAVs. In our previous experiments conducted using Epuck robots, the prediction radius is set to a fixed value of 0.08m, due to the relative poor perception ability of the robot \cite{cai2021behavior}. The robot calculates the concentration values of five equally distributed points on the circle with a radius 0.08m, and the point with the lowest concentration among these points is simply obtained as the desired destination of the next movement of the robot. However, defining motion direction in this way is very approximate, especially if the points selected are too few. In this work, we have chosen drones as the experimental platform and achieved a largely enhanced capability of visual perception.  We therefore can choose the points from five circles, starting from the one with a radius of 0.5m from the center of the rigid body of the agent to another four with an interval of a radius of 0.6m outwards. To consider the global optimality, the direction with the minimum sum of concentration values on the five circles is selected among the 180 directions (with an interval of interval 2°), and the next desired destination is obtained by adding the information of the motion step $s$. The vector is determined by:
\begin{equation}
\begin{split}
        \vec{v} = [k \cdot s \cdot\sin\theta,  k \cdot s \cdot\cos\theta, k \cdot \Delta h]
\end{split}
\end{equation}

Where $\theta$ is the angle between the desired direction and the Y-axis, $k$ is a scaling factor, and $\Delta h$ is the height difference relative to the observed target. Moreover, we use a dynamic $s$ to provide the UAV with a real-time adjusted maximum speed, which can accelerate the emergence of encirclement.

\subsection{Behavior Design of Swarm UAVs}

\begin{figure}[t]
        \centering
        \includegraphics[width=8.5cm]{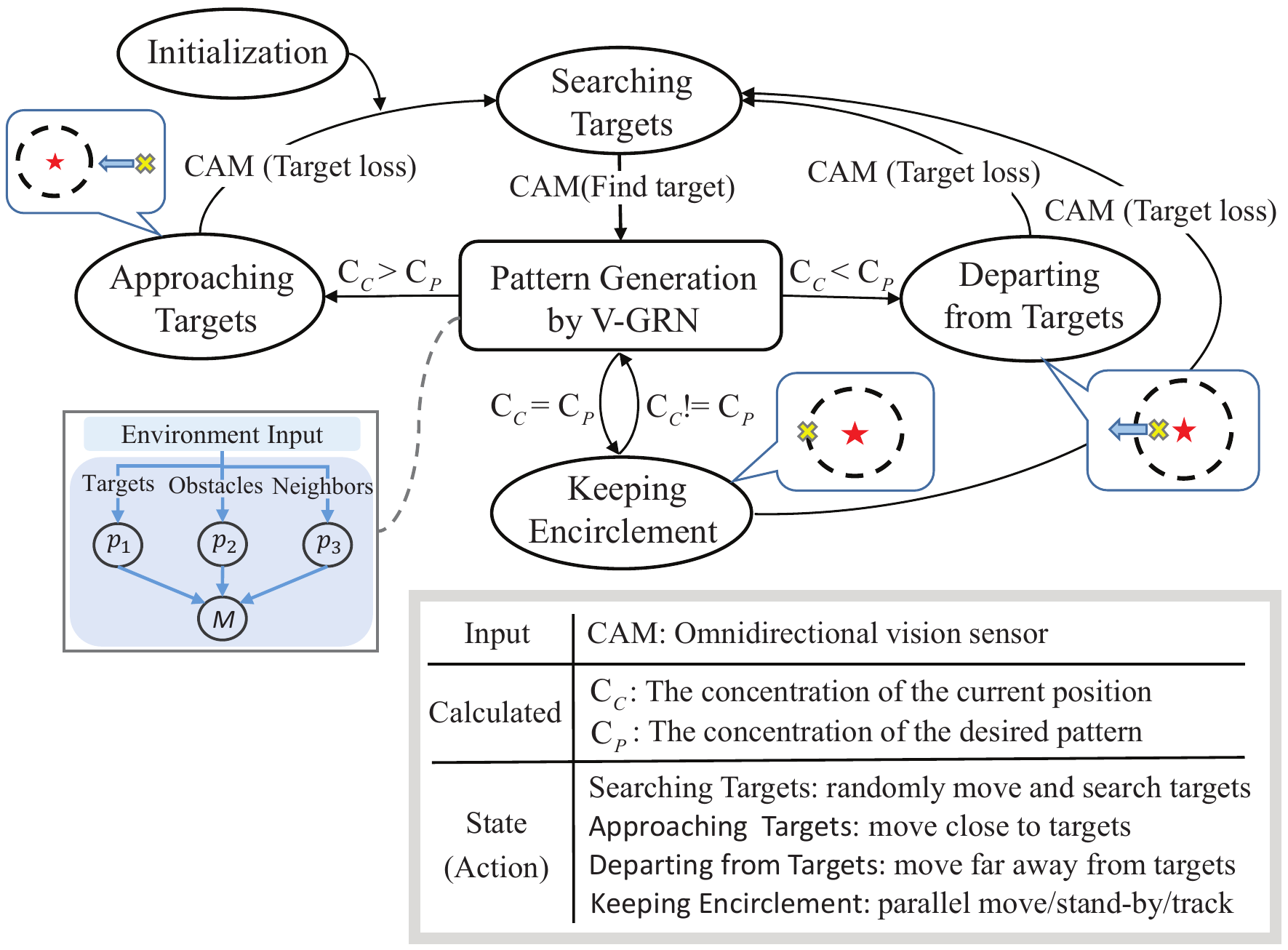}
        \caption{The design of the FSM model. The inputs include the positions of targets, obstacles, and neighbors, while the outputs are the behaviors of the UAV. The central part, “Pattern Generation by V-GRN”, involves a calculation process of the concentration field. The calculated $C_C$ and $C_P$ are used for defining triggering conditions.}
        \label{fig6}
\end{figure}

We designed the same FSM model (Fig. 6) for each individual UAV to maneuver in dynamic environments. The behavior of UAVs is decomposed into several parts: environmental perception, discovering targets, generating target entrapping patterns, and moving to the pattern point according to the generated concentration gradient. They can be encapsulated into four states, including searching targets, approaching targets, departing from targets, and keeping encirclement. Triggering conditions among states are also defined accordingly in the FSM model. 

Each UAV starts its task at "Initialization", then enters the "Searching Targets" state. At this state, the UAV moves randomly while searching for targets. It will keep this state until identifying at least one target. After discovering targets, the UAV starts generating the patterns by using V-GRN, and judges what to do next by using the calculated concentrations of the current location ($C_C$) and the desired pattern ($C_P$).  From here on, the UAV begins to track and entrap the targets. By comparing $C_C$ and $C_P$, it chooses the next state, including approaching targets, departing from targets, or keeping encirclement. In the state of keeping encirclement, to achieve stable encirclement, the UAVs can make corresponding motions according to the actions of the targets. If the targets stop moving, the UAVs will gradually reach a relatively stationary state. If the targets try to escape, the UAVs will parallelly move accordingly to form a new encirclement.
\section{EXPERIMENTAL VALIDATION}
\begin{figure}[b]
        \centering
        \includegraphics[width=8.5cm]{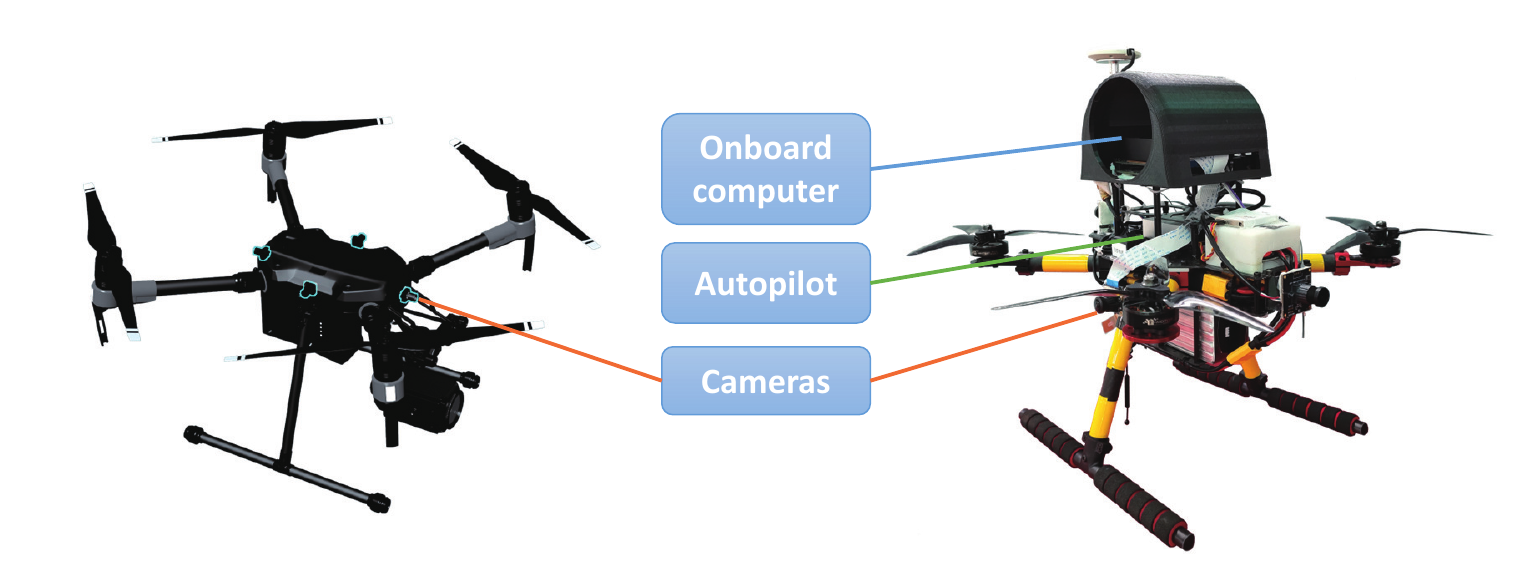}
        \caption{The left one is used in simulation experiments, while the right one is used in real-world experiments. They both have four cameras installed in four directions: front, back, left, and right.}
        \label{fig7}
\end{figure}
\subsection{Hardware Setup}
\subsubsection{Simulation setup}
To achieve a more realistic verification, we opt for AirSim \cite{airsim2017fsr} and the Unreal Engine 4 simulation platform for all simulation experiments. We replaced the default quadcopter model in AirSim with our customized quadcopter model, which provides four camera frames of front, back, left, and right to obtain surrounding environment information. We used the quadcopter as the captor and a military dual-rotor drone as the target. 

We built simulation environments and verified the proposed algorithm on two computers connected within a local area network, equipped with Intel i9-11900k and NVIDIA Quadro RTX 4000, one for UE4 simulation rendering and the other for running the proposed algorithm for all UAVs.

\subsubsection{Real-World setup}

We use a custom-built quadcopter named H350 as the captor and a DJI MATRICE 200 V2 quadcopter as the target. Each captor is equipped with four wide-angle RGB cameras providing a horizontal and vertical FOV of 120° and 90° respectively to obtain omnidirectional visual inputs. We acquired each camera frame in a resolution of 640×480 at a frequency of 20 Hz and stitched the four frames into one image. We mounted a Jetson Xavier NX with SSD for each captor and a DJI N3 as an autopilot (Fig. 7).

\subsection{Object Detection and Position Estimation}

Our customized network model has 269 layers and 1315729 parameters, with a size of only 3.0185 MB. Without being pretrained, this network is used to train our detector model on our simulation and real-world datasets, which include 1849 images with more than 5300 bounding boxes and 5567 training images with more than 5700 bounding boxes, respectively. 

In detection tasks, we set the IoU threshold to 0.5, the confidence threshold to 0.6, and the inference size to 1280 to detect the stitched image. Additionally, we use TensorRT to optimize the network model and record the perception speed of running YOLOv5m, YOLOv5s, and our network model on onboard devices.

We use the pose data provided by the UWB positioning system to calculate the visual positioning error. For each observation point, more than twenty visual position estimation data were taken to obtain the average localization error.

\subsection{Target Entrapping Experiments}

To verify that the proposed V-GRN method can guide the swarm of UAVs to emerge different entrapping patterns in different environments, we set up a series of simulation scenarios on AirSim simulator, including open, narrow road, and random obstacle scenarios.

The experimental process for the swarm of UAVs is as follows. The captors were placed behind the target at start, who constantly perceive their surroundings and search for targets using their own visual sensors independently. Once detecting one or more targets, the captors who observed the target will switch from the search state to the entrapment state. The captors will transform the entrapping patterns properly to avoid the obstacle when the safe distance condition is triggered. In the simulation experiments, we set the dynamic velocity range from 0 to 5m/s, while in the real-world experiments we limit the maximum speed to 1 m/s due to the large delay in acquiring raw image data.

We use the following quantitative evaluation metrics to evaluate the performance of the proposed method.
\begin{figure}[t]
        \centering
        \includegraphics[width=8.5cm]{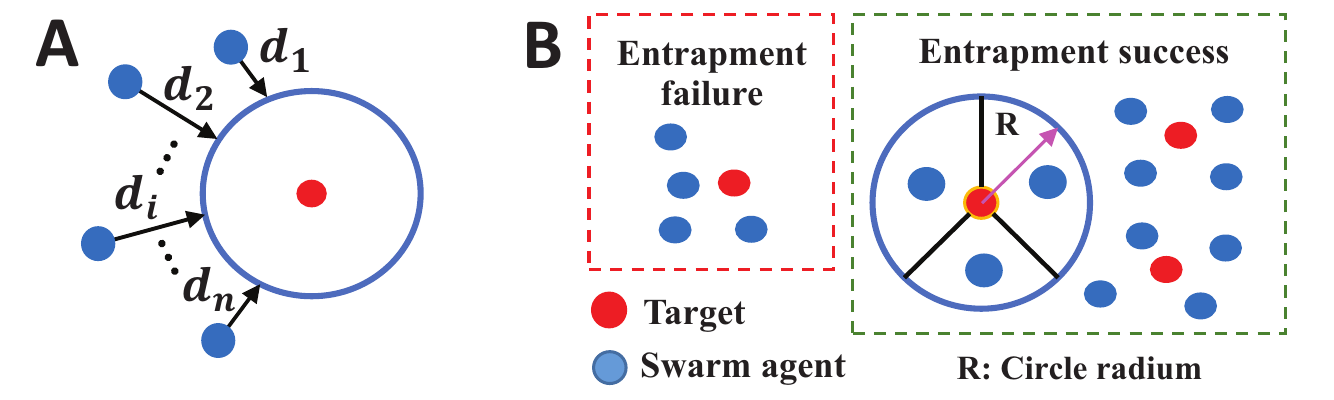}
        \caption{The schematic diagram of the entrapment distance error and the success rate. Part A shows the entrapment distance error ($d_i$) between agents and the encirclement circle. Part B shows the definition of the successful entrapment.}
        \label{fig8}
\end{figure}

\begin{itemize}

\item Average entrapment distance error (Fig. 8, part A): the average entrapment distance between the swarm of UAVs and the encirclement circle, which is calculated by:

\end{itemize}
\begin{equation}
\begin{split}
\overline{d} = \frac{\sum_{i = 1}^{n}{d_i}}{n} 
\end{split}
\end{equation}

\begin{itemize}

\item Average speed: The average speed of the captors from finding the targets to successful entrapment.
\item successful entrapment (Fig. 8, part B): We divide the circle area into three equal parts. Successful entrapment is defined to be achieved when more than one drone stops in each part.
\end{itemize}
\begin{figure}[b]
        \centering
        \includegraphics[width=8.2cm]{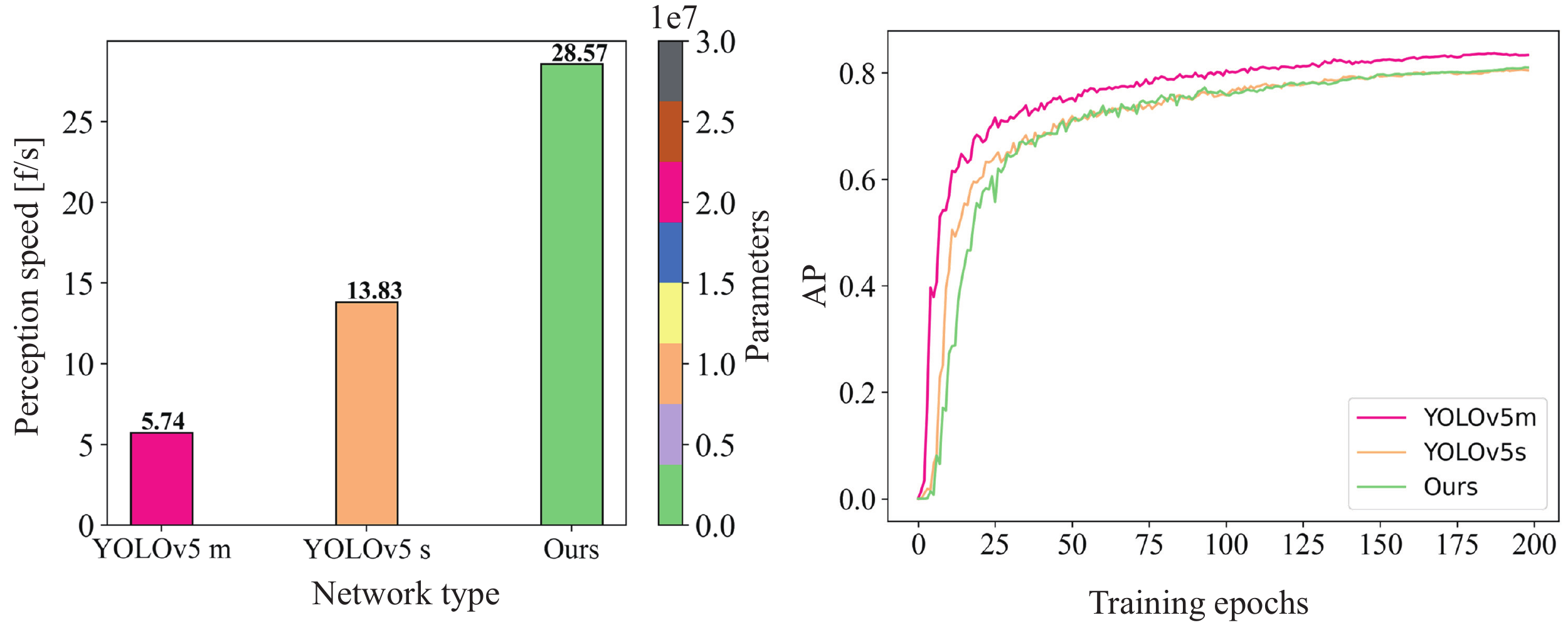}
        \caption{Results of the detector performance. We show model parameters and visual perception speed (left) tested in drone’s onboard computer and AP performance (right) for three type of Neural Networks: YOLOv5 m, YOLOv5 s and ours.}
        \label{fig9}
\end{figure}
\begin{figure*}[t]
        \centering
        \includegraphics[width=16.5cm]{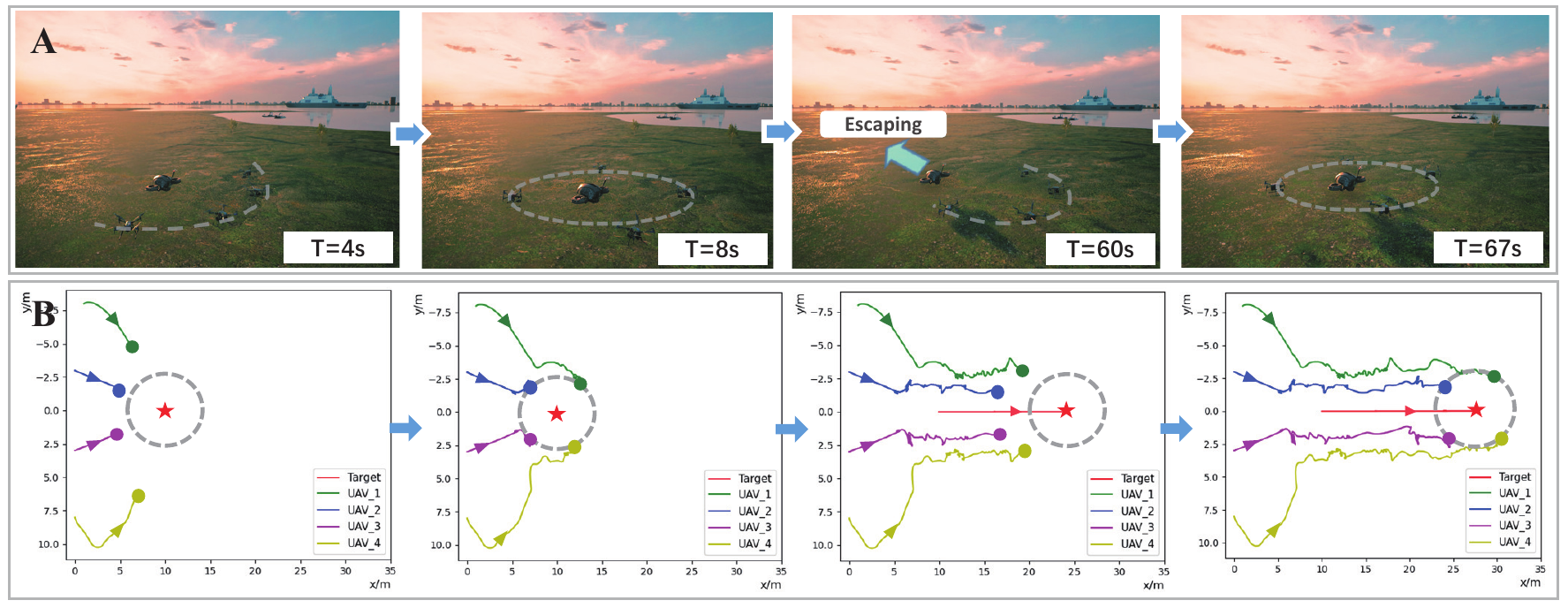}
        \caption{The entire process of UAVs swarm encircling a dynamic target. (A)  Experimental process. (B) The trajectory of all UAVs, and the NED coordinate system relative to the reference point is used. The corresponding entrapment distance error of all drones can be seen in Fig. 11.}
        \label{fig10}
\end{figure*}

\section{RESULT}
\subsection{Vision}

The prediction head we replaced is more sensitive to tiny instances. Therefore, the detector maintains high accuracy even if the number of convolution kernels is halved, and the results proved that the modifications we made substantially improved object detection speed with the same mean Average Precision as YOLOv5s. We present a training result of AP for 200 epochs (Fig. 13). The detector we deployed in real world experiment achieves a mean average precision of 97.6\%($AP_{50}$) and 80.46\%($AP$) at a confidence threshold of $p^{conf}$= 0.1\% on the hold-out valid set containing 650 images after 200 epochs of training.

The monocular visual relative localization error is mainly determined by the accuracy of bounding boxes, while the detector provides a slightly different bounding box information at different perspectives. Within the effective detection range (10 meters), the positioning estimation system presents a positioning error under 11\%.

\subsection{Swarm}

In the UE4 simulation, we validated the proposed algorithm through extensive experiments in three scenarios: open, narrow roads and random obstacle scenarios. The experimental results are as follows.
\begin{figure}[t]
	\centering
	\includegraphics[width=7.8cm]{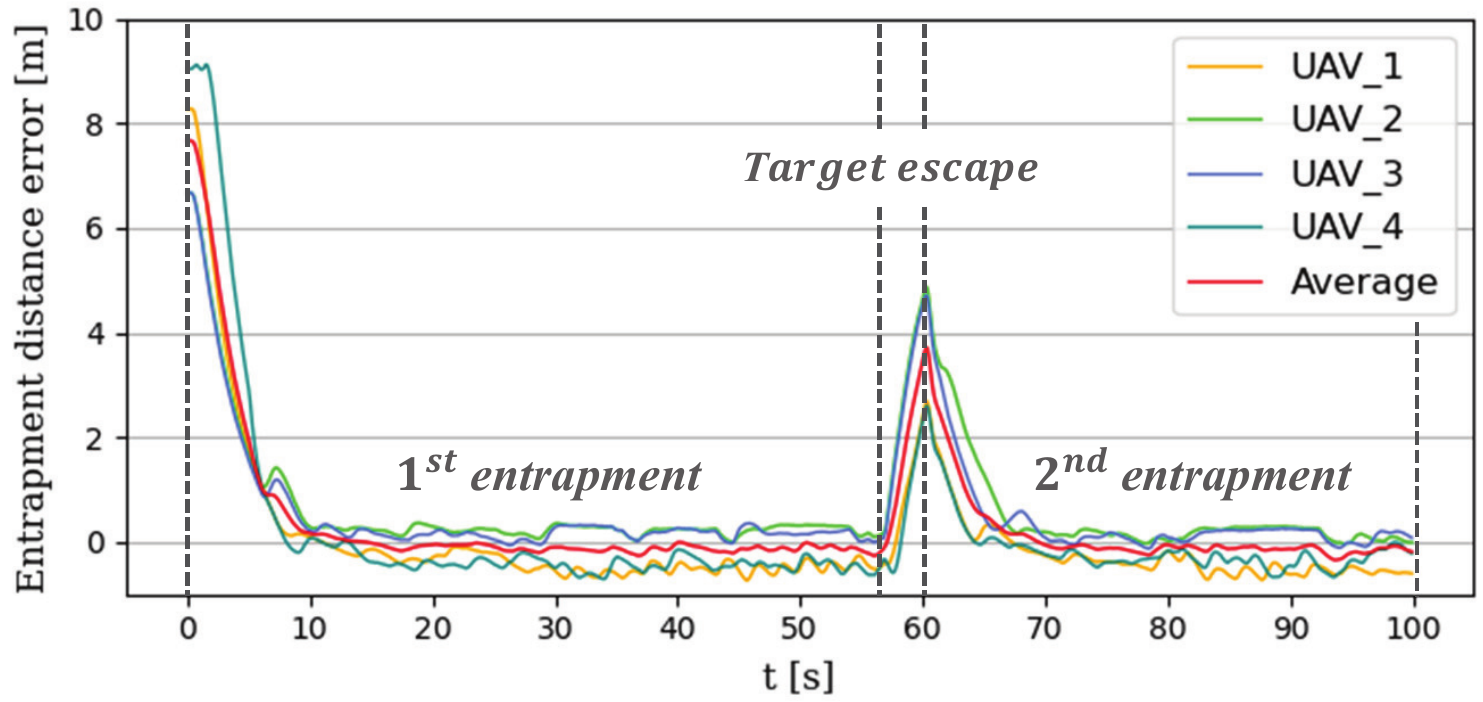}
	\caption{The entrapment distance error of UAVs. }
	\label{fig11}
\end{figure}
\begin{table}[b]
\centering
\caption{Experimental result}
\setlength{\tabcolsep}{3.6mm}{
        \begin{tabular}{ccccc}
        \toprule
        \multicolumn{1}{c}{\multirow{2}[4]{*}{ \shortstack{Initial average \\distance (m)}}} & \multicolumn{3}{c}{Success rate (\%)} & \multicolumn{1}{c}{\multirow{2}[4]{*}{\shortstack{Average speed\\(m/s)}}} \\
\cmidrule{2-4}          & 6s    & 10s   & 14s   &  \\
        \midrule
        6    & 0    & 93   & 100  & 0.76  \\
        10   & 80   & 100  & 100  & 2.01  \\
        14   & 73   & 100  & 100  & 2.53  \\
        \bottomrule
\end{tabular}
}
\label{tab1}
\end{table}
\subsubsection{In the open scenario}
Captors entrap the target at a dynamic speed ranging from 0.2m/s to 5m/s after spotting the target. From the results shown in Fig. 10, it can be seen that captors can quickly surround a static target more than 10 meters ahead within 10 seconds. Then, the target moved in a low speed and captors remained in a encirclement mode. After a few seconds the target started to escape from the encirclement in a high speed, the captors tried to catch up and entrap the target once again in a short period of time. Fig. 11 shows its entrapment distance error during the entire experimental process, which demonstrates the effectiveness of the proposed method in emerging entrapment behavior. We calculated the successful entrapment rates for more than 20 experiments at different average distances and different time consumption. The results can be seen in Table 1.
\subsubsection{In the narrow road scenario}
Captors first enter the parallel obstacle lanes in a encirclement mode (Fig. 12A), the concentration field around the target changes from a circular to an elliptical shape due to the observed obstacles. Captors maintain an elliptical pattern to dynamically entrap the target, guided by their own concentration field generated in real time by V-GRN. When the target and the captors enter the conical obstacle lanes (Fig. 12B), the concentration contour around the target presents a water droplet shape, with the formation of the captors changed accordingly. 
 
\begin{figure}[t]
        \centering
        \includegraphics[width=8.3cm]{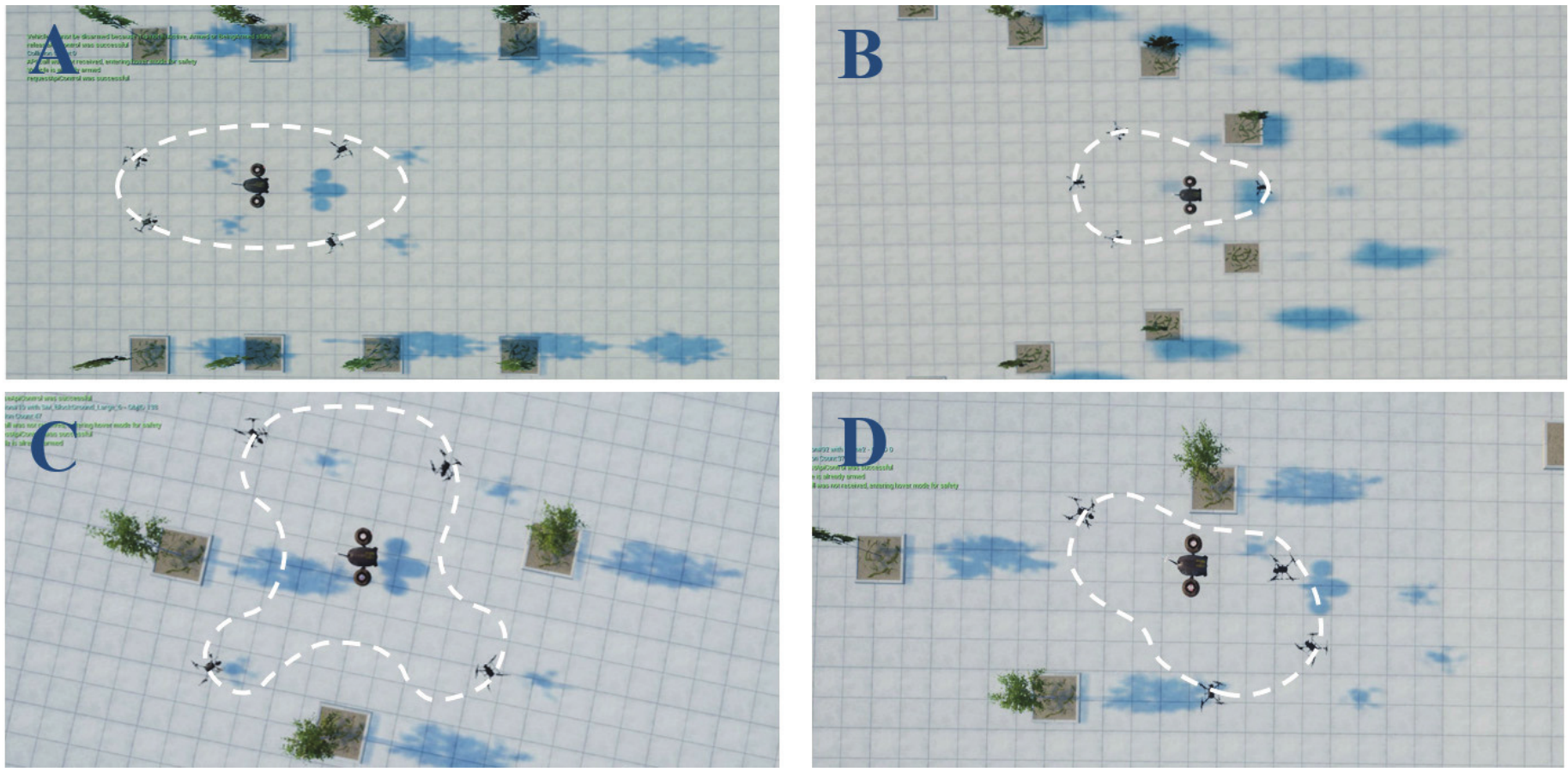}
        \caption{The entrapping pattern generated by the swarm of UAVs under different obstacle scenes. Among them, (A) is scene with parallel obstacle lanes. (B) is the scene with conical obstacle lanes. And (C, D) are the scenes with randomly distributed obstacles.}
        \label{fig12}
\end{figure}
\begin{figure}[t]
        \centering
        \includegraphics[width=8.5cm]{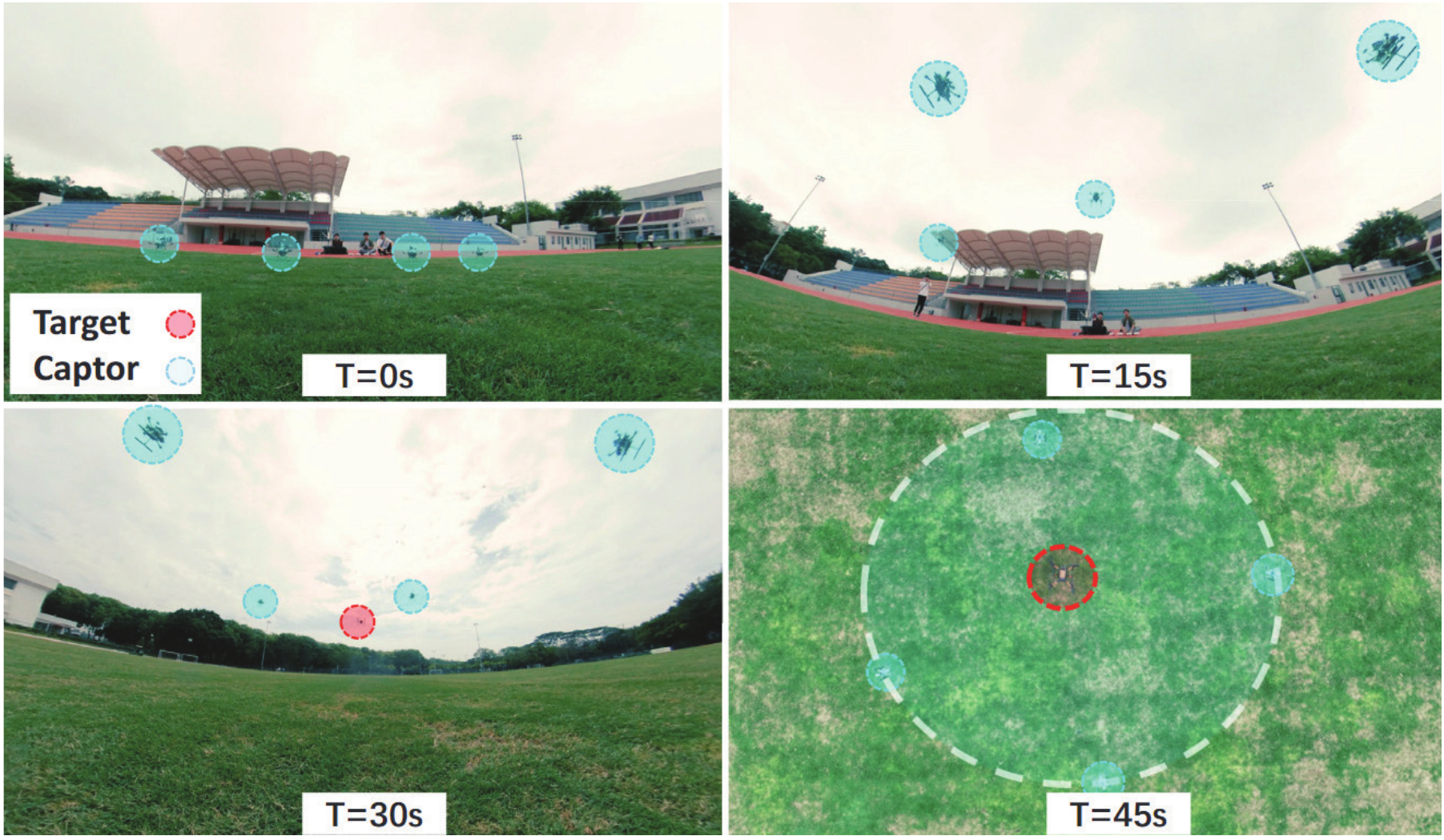}
        \caption{The processing of UAVs swarm behavior emergence in outdoor experiments(four UAVs entrap one target)}
        \label{fig13}
\end{figure}
\begin{figure}[t]
        \centering
        \includegraphics[width=8.5cm]{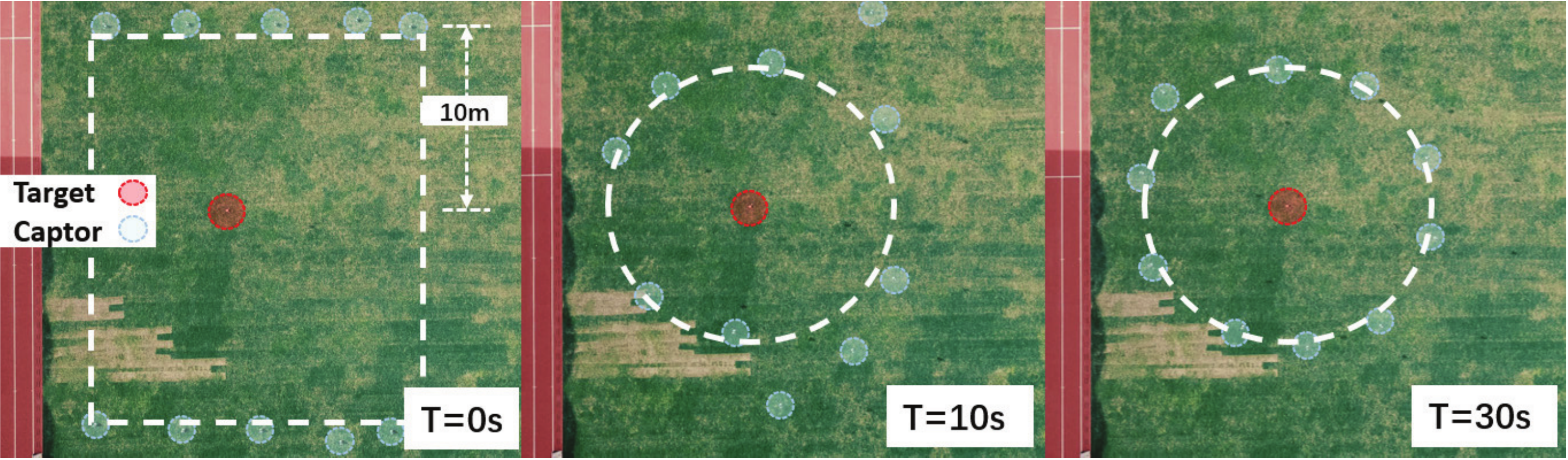}
        \caption{The processing of larger-scale UAVs swarm behavior emergence in outdoor experiments(Ten UAVs entrap one target).}
        \label{fig14}
\end{figure}
\subsubsection{In the random obstacle scenario}
When the captors and the target enter a scene with randomly distributed obstacles (Fig. 12(C, D)), the proposed V-GRN can still generate a variety of different concentration fields to guide the captors to switch to different entrapping patterns.

\subsubsection{Real-world experiments}
As showing in Fig. 13, four captors were placed 15 meters behind the target where they can not spot the target at first. While the start instruction was given at 10 seconds, these captors switched to searching state. At 20 seconds, the captors in front found the target and then switched state. The rest also completed state transition at 30s. The swarm of captors completed the entrapping task without collision at a average time of 50s. 

Besides, to illustrate that the proposed approach is still effective in large-scale UAV swarms, we increase the number of captors to 10. The result can be seen in Fig. 14. We also conducted fault-tolerant experiments, during which we operated some captors to land manually, to emulate the case when some drones may fail. The rest captors can still complete entrapping task with a success. 

The details of the above experiments can be seen in the online video: \href{https://youtu.be/nYnYlolmLiI}{https://youtu.be/nYnYlolmLiI} 

\section{CONCLUSIONS AND FUTURE WORK}
We proposed a vision-based gene regulatory network and object position estimation algorithm that enables multiple UAVs to complete the target entrapment task and emerge intelligent swarm behavior without collision in dynamic and complex environments. The proposed method does not rely on inter-agent communication and external location information, which can work well in GNSS-denied environments or in situations where the existing communications are unreliable. Our simulation and real-world experiments demonstrate that the proposed approach is robust and effective, which can resist partial agent failures and still fulfill the entrapping task. It is notable that we increased the number of captors to 10 in real-world entrapping experiments, which to our best knowledge is the first work reported in this line. 

Our work paves the way for practical application of a swarm of UAVs in vision-based encirclement. As a seminal work, the proposed work needs improvements in several directions. For instances, the monocular position estimation algorithm can be improved so that the system can estimate the position even if the object is not detected in the vision, and the time consuming to read and process multiple camera frames can be reduced by hardware to enable a swarm of UAVs to fly in a higher speed.
\bibliographystyle{IEEEtran}
\bibliography{IEEEabrv, root.bib}

\end{document}